\documentclass{article}
\usepackage{iclr2026_conference,times}
\usepackage[T1]{fontenc}
\usepackage[utf8]{inputenc}
\usepackage{amsmath,amssymb}
\usepackage{microtype}
\usepackage{graphicx}
\usepackage{booktabs,tabularx,array,colortbl}
\usepackage{fvextra}
\usepackage{enumitem}
\usepackage{caption}
\usepackage{placeins}
\usepackage{needspace}
\usepackage{xurl}
\usepackage{hyperref}
\usepackage{bookmark}

\definecolor{sodablue}{RGB}{40,78,116}
\definecolor{sodarule}{RGB}{210,221,230}
\hypersetup{colorlinks=true,linkcolor=sodablue,citecolor=sodablue,urlcolor=sodablue,
  pdftitle={SciWalker: Synthesizing Scientific Coding Problems with Operator Graphs and Execution Feedback},
  pdfauthor={Chenxi Li and Yun Luo},
  pdfsubject={Scientific data generation with operator guidance and execution feedback}}
\renewcommand{\tabularxcolumn}[1]{>{\raggedright\arraybackslash}p{#1}}
\newcolumntype{R}[1]{>{\raggedleft\arraybackslash}p{#1}}
\newcolumntype{G}{>{\columncolor{sodablue!8}[\tabcolsep][0pt]\raggedleft\arraybackslash}p{0.13\linewidth}}
\setlist[itemize]{leftmargin=1.4em,itemsep=2pt,topsep=4pt}

\DeclareUnicodeCharacter{2192}{\ensuremath{\rightarrow}}
\DeclareUnicodeCharacter{00D7}{\ensuremath{\times}}

\newcommand{\SWPipelineWidth}{1.0\linewidth}

\title{SciWalker: Synthesizing Scientific Coding Problems\\with Operator Graphs and Execution Feedback}
\author{%
\begin{minipage}[t]{\dimexpr\textwidth-2\tabcolsep\relax}
\centering\normalfont
\textbf{Chenxi Li}\textsuperscript{1,2}\hspace{1em}%
\textbf{Wenxuan Zeng}\textsuperscript{2,3}\hspace{1em}
\textbf{Yun Luo}\textsuperscript{2\textdagger,$\ddagger$}\hspace{1em}
\textbf{Fangchen Yu}\textsuperscript{2}\hspace{1em}\\ 
\textbf{Peng Ye}\textsuperscript{2}\hspace{1em} 
\textbf{Yu Cheng}\textsuperscript{4}\hspace{1em} 
\textbf{Jun Zhang}\textsuperscript{1$\ddagger$}\hspace{1em} 
\\[0.6em]
{\small
\textsuperscript{1} The Hong Kong University of Science and Technology \quad
\textsuperscript{2} Shanghai AI Laboratory\\[0.15em]
\textsuperscript{3} Tsinghua University \quad
\textsuperscript{4} Nanyang Technological University
\\[0.3em]
\textsuperscript{\textdagger}Project Lead,$\quad$ \textsuperscript{$\ddagger$}Corresponding authors
}
\end{minipage}}
\iclrfinalcopy

\begin{document}
\raggedbottom
\maketitle
\fancyhead[L]{\small SciWalker: Synthesizing Scientific Coding Problems with Operator Graphs and Execution Feedback}
\fancyhead[R]{}
\begin{abstract}
Improving the scientific coding capabilities of large language models (LLMs) requires high-quality training data. However, such data remain scarce because manually authoring realistic problems is costly and time-consuming, while systematically covering diverse scientific domains and algorithmic combinations remains challenging. To address this, we introduce SciWalker, a framework for synthesizing scientific coding problems through operator-chain sampling and execution feedback. The framework combines scientific library interfaces with operation modes to instantiate operators, organizes them into operator graphs, and samples operator chains as computational workflow cues. Guided by these cues, we adopt LLMs to generate scientifically grounded problem statements, reference solutions, and tests, with failed generations iteratively repaired using execution feedback. By combining structured workflow composition with verification and quality review, SciWalker enables scalable task generation while promoting scientific grounding, computational diversity, and executability. Using this framework, we construct 8,178 high-quality problems spanning 5 scientific domains and 32 subdomains. To evaluate their training utility, we conduct reinforcement learning on Qwen3.5-9B using the GSPO algorithm. This training improves SciCode subproblem accuracy by 9.9 percentage points, from 29.3\% to 39.2\%, with gains across scientific code generation, code repair, and reasoning benchmarks.
The code for SciWalker is available at \url{https://github.com/lichenx1/SciWalker}.

\end{abstract}

\section{Introduction}
\label{sec:1}



Scientific code plays a crucial role in translating scientific theories into computational practice across a wide range of disciplines~\citep{2020SciPy-NMeth}. High-quality, diverse training data have been shown to improve the scientific reasoning and code generation capabilities of large language models (LLMs)~\citep{zhang2024sciinstruct,wei2024magicoder}.
However, such data remain scarce because constructing realistic scientific coding problems typically requires substantial expert effort to formulate scientific contexts, compose computational procedures, implement reference solutions, and design reliable tests~\citep{tian2024scicode}. Manually carrying out these steps is costly and time-consuming~\citep{lai2023ds1000}, while systematically covering diverse scientific domains, subdomains, and algorithmic combinations remains challenging.

To address these challenges, we introduce SciWalker, a framework for scalable synthesis of multistep scientific coding problems. Our central idea is to leverage scientific computing libraries as computational building blocks and compose their interfaces into diverse scientific workflows as illustrated in Figure \ref{fig:pipeline}. Specifically, SciWalker combines library interfaces with operation modes to instantiate operators, organizes them into operator graphs, and samples operator chains as workflow cues. Guided by these cues and corresponding scientific contexts, LLMs generate problem statements, reference solutions, and tests with well-defined computational objectives. Execution feedback is then used to identify and iteratively repair failures in the generated code and tests, while quality review further filters the resulting problems. 

Based on SciWalker, we construct a large-scale dataset of 8,178 scientific coding problems spanning five major domains, including mathematics, physics, chemistry, biology, and materials science. The problems are generated from operator chains of varying lengths and subsequently refined through execution-guided repair and quality review, resulting in diverse computational workflows grounded in realistic scientific contexts. Collectively, the dataset contains 37,820 computational substeps, with each problem integrating multiple interdependent operations rather than isolated function calls.

To assess the training value of the generated data, we conduct reinforcement learning on Qwen3.5-9B~\citep{qwen3.5} using step-level execution rewards and GSPO policy updates~\citep{zheng2025gspo}. This training improves SciCode subproblem accuracy from 29.3\% to 39.2\%, a gain of 9.9 percentage points over the baseline. Gains also extend to scientific code generation, code repair, and reasoning benchmarks, with improvements of 17.3 percentage points on DS-1000 and 6.0 percentage points on MATH-500.

The main contributions of our work are as follows:

\begin{figure}[t]
\centering
\includegraphics[width=\SWPipelineWidth,keepaspectratio]{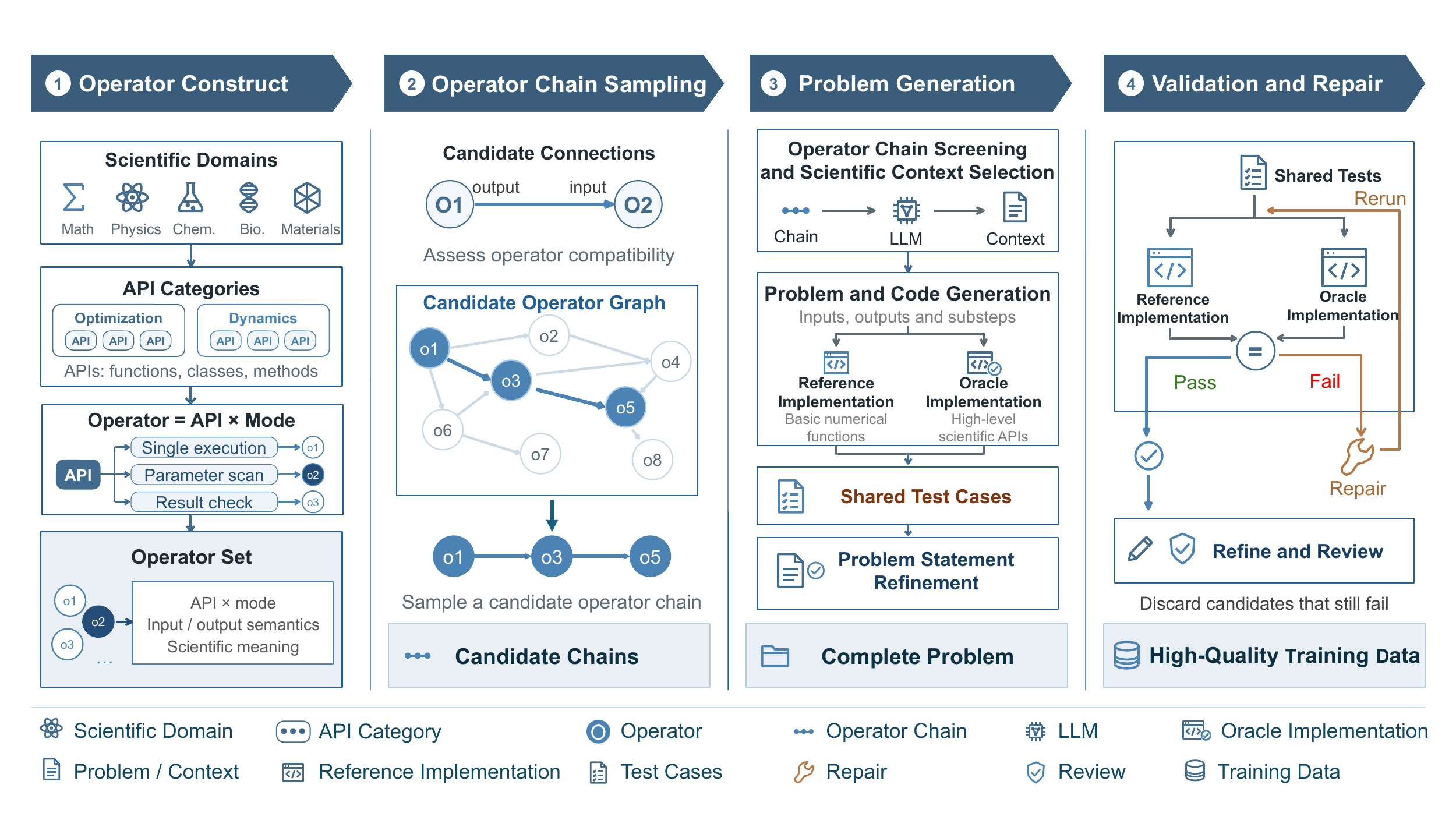}
\caption{Scientific code training data generation workflow. Scientific library APIs are combined with operation modes to form operators, which are used to construct a graph and sample candidate operator chains. Large language models screen these chains and select scientific contexts, generate tasks, reference implementations, and tests, and then refine the problem statements accordingly. Execution validation, feedback-driven repair, and quality review yield high-quality training data.}
\label{fig:pipeline}
\end{figure}

\begin{itemize}
\item
  We propose SciWalker, which combines scientific operator composition with execution feedback to guide large language models in automatically generating and revising multistep scientific coding problems.
\item
  We construct 8,178 high-quality scientific problems covering 5 scientific domains and 32 subdomains, accompanied by reference solutions and tests.
\item
  We conduct reinforcement learning experiments to assess the training value of the generated data. The trained model shows improvements on both in-domain scientific coding tasks and out-of-domain code repair and reasoning benchmarks.
\end{itemize}

\FloatBarrier

\section{Related Work}
\label{sec:2}

\textbf{Scientific and code data construction.} {SciCode~\citep{tian2024scicode}} provides real research problems, reference solutions, and tests curated by scientists to evaluate multistep scientific coding capabilities. Automated data construction approaches draw on existing problems, scientific literature, and code resources. {SciInstruct~\citep{zhang2024sciinstruct}} augments existing scientific problems with model-generated reasoning, refined through self-review and revision. {WildSci~\citep{liu2026wildsci}} synthesizes scientific multiple-choice questions from peer-reviewed literature, supporting reinforcement learning through unambiguous answer evaluation. {UniScientist~\citep{li2026uniscientist}} combines validated scientific claims with evidence retrieval to generate open-ended research questions, accompanied by rubrics revised and validated by language models and domain experts. {Magicoder~\citep{wei2024magicoder}} uses open-source code snippets to guide the generation of programming tasks. {OpenCodeInterpreter~\citep{zheng-etal-2024-opencodeinterpreter}} trains code models on multi-turn interactions incorporating execution and human feedback for iterative refinement. Scaling multistep scientific coding data requires diverse combinations of scientific computations, coherent tasks with clear scientific objectives, and reference solutions with tests. Our work addresses these requirements through scientific operator composition and scientific context design, using execution feedback to repair and select the generated problems.



\section{Synthesizing Scientific Coding Problems with Operator Graphs and Execution Feedback
}
\label{sec:3}

\begin{figure}[t]
\centering
\includegraphics[width=\linewidth]{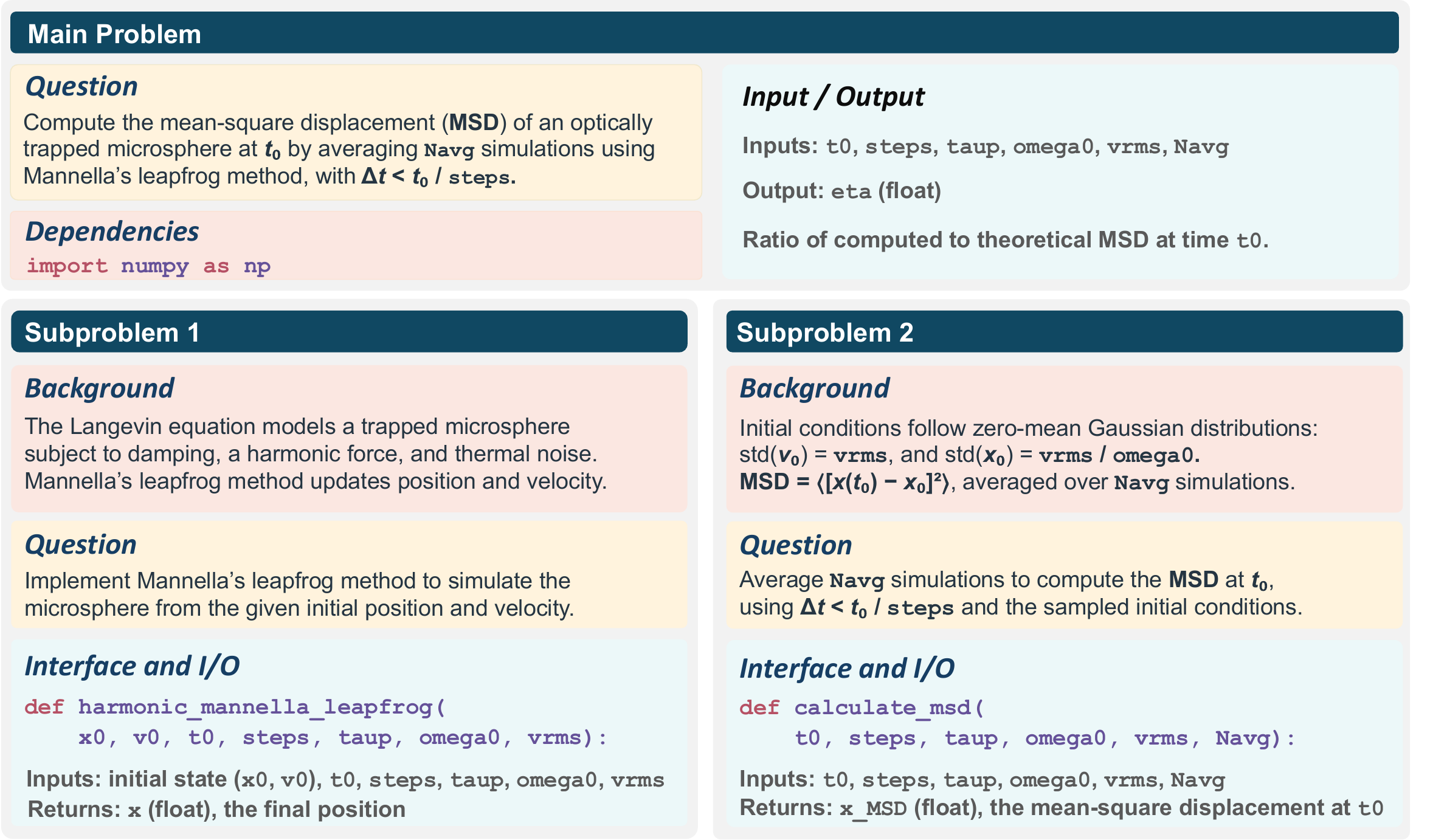}
\caption{Multi-turn scientific coding problem format.}
\label{fig:problem_format}
\end{figure}

Scientific coding tasks often involve interdependent computations, where later steps build on functions implemented earlier. We therefore adopt a multi-turn format that preserves these dependencies and supports stepwise validation. Following SciCode~\citep{tian2024scicode}, each problem consists of a main problem and an ordered sequence of subproblems, as illustrated in Figure~\ref{fig:problem_format}. The main problem defines the overall objective and dependencies. Each subproblem provides a task description, background, and interface specification. At each turn, the model implements the current subproblem using its specification and previously generated code.

Our framework combines LLMs with programmatic workflows to construct scientific problems containing problem statements, reference solutions, and tests. 
The overall workflow consists of four stages as illustrated in Figure \ref{fig:pipeline}: (1) Scientific operator construction; (2) Operator graph sampling; (3) Scientific problem generation; (4) Validation, Repair, and Final refinement.

\subsection{Operator Construction}
\label{sec:3.1} 
We first organize scientific library interfaces by computational purpose into categories such as optimization, dynamics, and matrix decomposition. Each API, defined as a function, class, or method, is combined with operation modes such as single-case execution, parameter sweeps, and result checking. We define each API $\times$ mode combination as an operator, which serves as a node for graph construction and operator chain sampling. Representing different uses of the same API as distinct operators expands a finite collection of APIs into a larger operator space, enabling more diverse computational workflows for scientific problem generation.


\subsection{Operator Graph Sampling}
\label{sec:3.2}
Each node in the operator graph represents an API × mode combination, as defined in Section~\ref{sec:3.1}. Within each subdomain, the framework scores directed connections between operators according to their input/output compatibility and operational workflows, retaining multiple high-scoring successors for each node. Operator chains are then sampled through random walks. Starting from a selected operator, the framework repeatedly chooses among its retained successors until the target chain length is reached. Varying the starting node and walk path produces diverse operator combinations from the same graph. The resulting chains and their node descriptions serve as computational workflow cues for subsequent scientific problem design.


\subsection{Scientific Problem Generation}
\label{sec:3.3}
Given a sampled operator chain, the framework generates a problem through three stages: (1) \textbf{Chain assessment}, it evaluates whether the chain can support a coherent computational workflow with a natural scientific context, well-defined target quantities, and meaningful multistep reasoning. (2) \textbf{Context construction}, the framework specifies the input conditions, solution objectives, and roles of individual operators, followed by an independent plausibility review. Rejected contexts are regenerated and reviewed within a limited number of rounds. (3) \textbf{Problem authoring}, using the approved context and operator chain, the framework generates the scientific task and substeps, tests, a reference implementation based on basic numerical functions, and an oracle implementation using high-level scientific APIs. The problem statement is then revised for consistency with the generated code and tests. 

The final problem contains the main problem, subproblems, input/output specifications, reference solutions, and tests. Substeps are organized according to the scientific task and its dependencies, after which the problem proceeds to execution validation and repair.


\begin{figure}[t]
\centering
\includegraphics[width=\linewidth,keepaspectratio]{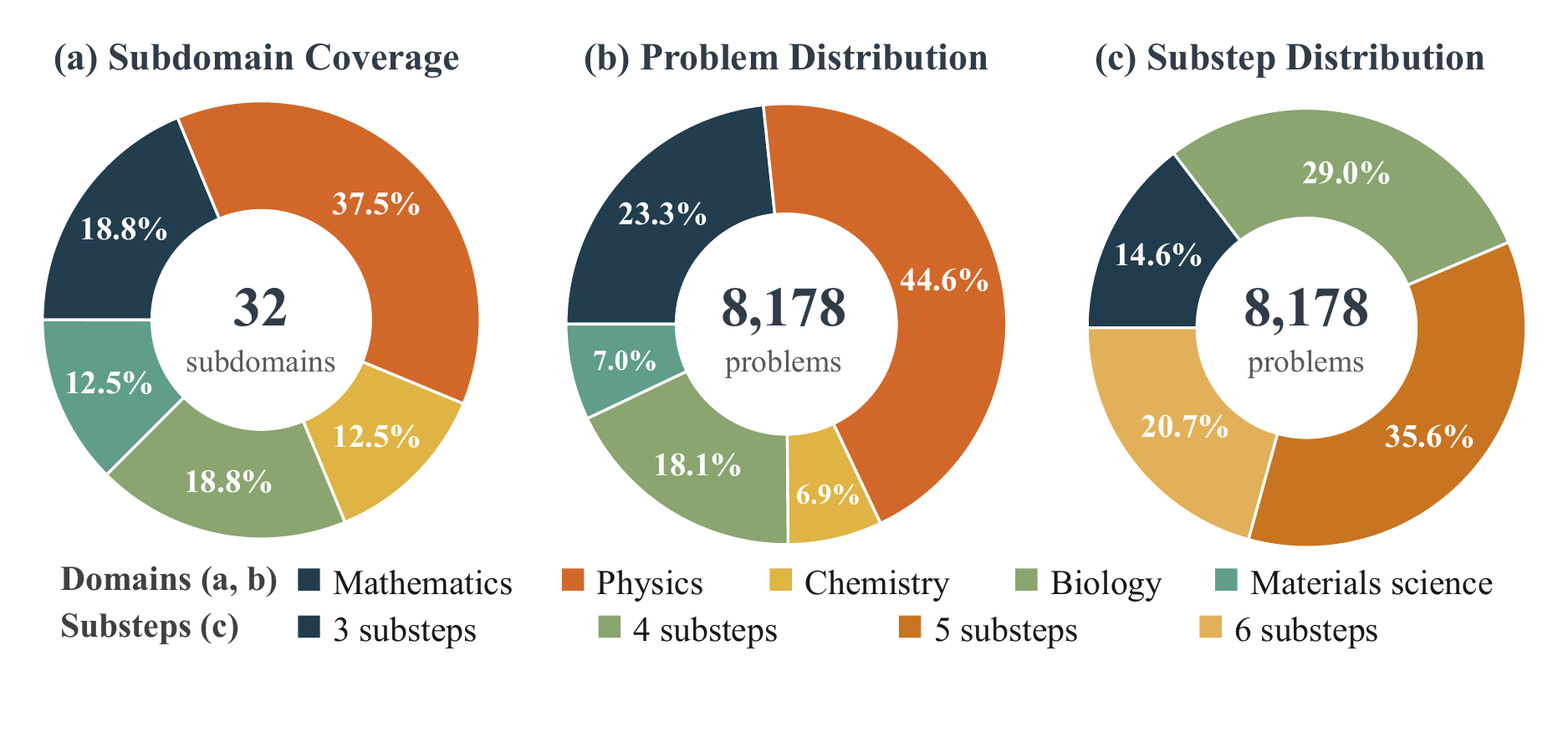}
\caption{Dataset coverage and composition. (a) Domain distribution of 32 subdomains. (b) Domain distribution of 8,178 high-quality problems. (c) Distribution of substep counts across the same 8,178 problems. Sector areas represent category proportions, with percentages labeled on the sectors and categories identified in the legends.}
\label{fig:dataset_statistics}
\label{fig:subdomains}
\label{fig:domains}
\label{fig:substeps}
\end{figure}

\subsection{Validation, Repair, and Final Refinement}
\label{sec:3.4}

After a problem is generated, it undergoes three stages to ensure its executability and overall quality: (1) \textbf{Execution validation}, the framework checks the problem structure, interfaces, and dependencies, executes substep and end-to-end tests, and performs differential testing between the reference and oracle implementations. The tests cover numerical correctness, parameter variations, boundary cases, and scientific invariants. (2) \textbf{Execution-guided repair}, when validation fails, error messages are fed back to the model to repair the problem statement, tests, or implementations. Each revision is revalidated, and candidates that exceed the repair limit are discarded. (3) \textbf{Quality refinement}, validated problems are polished and assessed for scientific naturalness, task difficulty, implementation complexity, and test strength. Candidates below the quality threshold undergo limited rounds of refinement and renewed validation, after which only qualified problems are retained.





\subsection{SciWalker Implementation and Dataset Statistics}
\label{sec:3.5}

We use DeepSeek-V4-Flash~\citep{deepseekai2026deepseekv4} throughout the entire data generation pipeline. With stage-specific prompts, the same model performs chain review, context selection, problem generation, execution-guided repair, polishing, and quality review. We enable its maximum reasoning setting and use a context window of 200,000 tokens.

We organize APIs from scientific Python libraries (Appendix~\ref{app:e}) into subdomain-specific operator catalogs, which serve as the basis for our data generation pipeline. The resulting problems span five broad scientific domains—mathematics, physics, chemistry, biology, and materials science, covering 32 subdomains. Figure~\ref{fig:dataset_statistics}(a) shows the distribution of these subdomains across the five domains. For each subdomain, we construct 560 operators, yielding 17,920 operators and 772,178 graph edges in total. Operator chains of lengths 3-15 are then sampled as workflow cues for problem generation.

After execution-based validation and quality filtering, we retain 8,178 high-quality problems, whose domain distribution is shown in Figure~\ref{fig:dataset_statistics}(b). Appendix~\ref{app:b} reports candidate retention across the generation stages. Each problem comprises 3--6 substeps, resulting in 37,820 substeps overall, with a mean of 4.62 and a median of 5 per problem; the corresponding distribution is shown in Figure~\ref{fig:dataset_statistics}(c). The number of substeps does not directly correspond to operator-chain length. As discussed in Section~\ref{sec:3.3}, operator chains provide high-level computational workflow cues, while the model structures the final substeps according to the scientific objective and computational dependencies of each problem.

\section{Reinforcement Learning for Scientific Coding}
\label{sec:4}

To evaluate the effectiveness of the data generated by SciWalker for training scientific coding models, we perform reinforcement learning with execution-based rewards. We use GSPO for policy optimization with substep execution rewards.
\subsection{Policy Optimization with GSPO}
\label{sec:4.1}

We optimize the policy with Group Sequence Policy Optimization (GSPO), using the GSPO-token formulation introduced by \citet{zheng2025gspo}. For each main problem, we sample multiple complete solution trajectories and expand them into substep-level prompt--response samples. Each prompt contains the current task and preceding code from the same trajectory. The response contains the current substep's reasoning and final code. Only the generated response tokens contribute to the loss.
Responses to the same substep of the same main problem form a group, although their preceding code may differ. Within each group, the advantage of trajectory \(i\) at substep \(t\) is
\begin{equation}
\label{eq:grpo-advantage}
A_{i,t}=\frac{r_{i,t}-\mu_t}{\sigma_t+\delta},
\end{equation}
where \(r_{i,t}\) is the training reward defined in Equation~\ref{eq:execution-reward}, \(\mu_t\) and \(\sigma_t\) are the group reward mean and sample standard deviation, and \(\delta\) is a small constant for numerical stability.

For policy optimization, we denote a substep sample by \((x_i,y_i,A_i)\), where \(y_i\) is the complete response for one substep. GSPO uses the length-normalized sequence probability ratio
\begin{equation}
\label{eq:sequence-ratio}
\rho_i(\theta)=
\left(\frac{\pi_\theta(y_i\mid x_i)}{\pi_{\mathrm{old}}(y_i\mid x_i)}\right)^{1/|y_i|},
\end{equation}
where \(\pi_{\mathrm{old}}\) is the fixed old policy. GSPO-token uses
\begin{equation}
\label{eq:token-ratio}
\widehat\rho_{i,k}(\theta)=
\operatorname{sg}\!\left[\rho_i(\theta)\right]
\frac{\pi_\theta(y_{i,k}\mid x_i,y_{i,<k})}
{\operatorname{sg}\!\left[\pi_\theta(y_{i,k}\mid x_i,y_{i,<k})\right]},
\end{equation}
where \(k\) indexes response tokens and \(\operatorname{sg}\) stops gradient propagation. Every token shares the sequence ratio's forward value, while gradients propagate through its own log probability. The GSPO-token objective is
\begin{equation}
\label{eq:gspo-objective}
\mathcal J_{\mathrm{GSPO\text{-}token}}(\theta)=
\frac{1}{|\mathcal B|}\sum_{i\in\mathcal B}
\frac{1}{|y_i|}\sum_{k=1}^{|y_i|}
\min\!\left[
\widehat\rho_{i,k}(\theta)A_i,\,
\operatorname{clip}\!\left(\widehat\rho_{i,k}(\theta),1-\epsilon_{\mathrm{low}},1+\epsilon_{\mathrm{high}}\right)A_i
\right],
\end{equation}
where \(\mathcal B\) denotes a batch of substep training samples, and \(\epsilon_{\mathrm{low}}\) and \(\epsilon_{\mathrm{high}}\) specify the clipping range.

All tokens in a substep response share the same advantage \(A_i\), so this formulation is equivalent to GSPO in objective value, clipping conditions, and gradient. Additional rollout probability correction and numerical safeguards used in training are detailed in Appendix~\ref{app:d}.

\subsection{Constructing Substep Training Samples}
\label{sec:4.2}

After complete trajectories are generated and scored, they are expanded into training samples by substep. Let main problem \(q\) have \(T_q\) steps requiring model responses. Each trajectory then yields \(T_q\) prompt--response samples, and eight trajectories yield \(8T_q\) samples in total. The sample for trajectory \(i\) at step \(t\) is denoted by \((x_{q,i,t},y_{q,i,t})\), where \(x_{q,i,t}\) contains the current task, interface requirements, and preceding code from the same trajectory, while \(y_{q,i,t}\) contains the reasoning and final code generated for the current step.

Each sample is optimized only over the valid generated tokens in its current response, with both reasoning and the final code contributing to the loss. Preceding code belongs to the prompt and is not counted again in the loss as a prediction target for the current step. Each substep retains its own reward, and rewards from subsequent steps are not accumulated backward.


Each substep receives an execution reward of 1 if all its tests pass and 0 otherwise:
\begin{equation}
\label{eq:execution-reward}
r_{i,t}=\mathbf{1}\!\left[\text{all tests for substep }t\text{ pass}\right].
\end{equation}




\FloatBarrier

\begin{table}[t]
\centering
\caption{Shared reinforcement learning and evaluation settings.}
\label{tab:training_settings}
\small
\setlength{\tabcolsep}{4pt}
\renewcommand{\arraystretch}{1.17}
\begin{tabularx}{\linewidth}{@{}>{\raggedright\arraybackslash}p{0.29\linewidth}X@{}}
\toprule
\textbf{Item} & \textbf{Shared setting} \\
\midrule
Learning rate & \(1\times10^{-6}\) \\
Optimizer parameters & Adam, \(\beta_1=0.9\), \(\beta_2=0.98\), weight decay \(=0.1\) \\
GSPO clipping & \(\epsilon_{\mathrm{low}}=3\times10^{-4}\), \(\epsilon_{\mathrm{high}}=4\times10^{-4}\) \\
Gradient clipping & 1.0 \\
KL / entropy terms & KL reward, KL loss, and entropy regularization disabled \\
Training budget & 150 steps, 32 problems per step, 8 trajectories per problem \\
Training sampling & temperature=1.0, top\_p=1.0, top\_k=-1 \\
Training input / generation limit & 40,000 / 20,000 tokens \\
Training subproblem execution timeout & 120 seconds \\
Evaluation settings & Thinking, generation limit of 32,768 tokens, execution timeout of 300 seconds \\
Evaluation sampling & temperature=0.6, top\_p=0.95, top\_k=20 \\
\bottomrule
\end{tabularx}
\end{table}

\section{Experiments}
\label{sec:5}


\subsection{RL Training and Evaluation Settings}
\label{sec:5.1}

The shared training and evaluation settings are summarized in Table~\ref{tab:training_settings}. As an additional baseline, we apply PPO-based RL with a 50-step warm-up schedule.

\textbf{Evaluation}. 
We adopt 16 different evaluation benchmarks to evaluate the training effectiveness, which comprise five benchmarks for scientific computing and code generation, six for code repair and execution, and five for reasoning and knowledge:
\begin{enumerate}
    \item Scientific computing and code generation: SciCode~\citep{tian2024scicode}, DS-1000~\citep{lai2023ds1000}, HumanEval~\citep{chen2021codex}, LiveCodeBench Code Generation v6~\citep{jain2025livecodebench}, and APPS Introductory~\citep{hendrycks2021apps};
    \item Code repair and execution: HumanEvalFix-Python, HumanEvalFix-Rust, and HumanEvalFix-JavaScript~\citep{muennighoff2024octopack}, QuixBugs-Java and QuixBugs-Python~\citep{lin2017quixbugs}, and LiveCodeBench Execution v2~\citep{jain2025livecodebench};
    \item Reasoning and knowledge: MATH-500~\citep{lightman2024letsverify}, GSM8K~\citep{cobbe2021gsm8k}, BBH Multistep Arithmetic~\citep{suzgun2023bbh}, ARC-Easy~\citep{clark2018arc}, and MMLU-Pro Computer Science~\citep{wang2024mmlupro}.
\end{enumerate}
Each trained model uses a fixed checkpoint across benchmarks, and the best checkpoint is selected via the evaluation score of SciCode.
Across the 16 benchmarks, we calculate the unweighted mean of the benchmark scores, each averaged over three evaluations.

\subsection{Results Comparison}
\label{sec:5.2}

\begin{figure}[t]
\centering
\includegraphics[width=\linewidth,keepaspectratio]{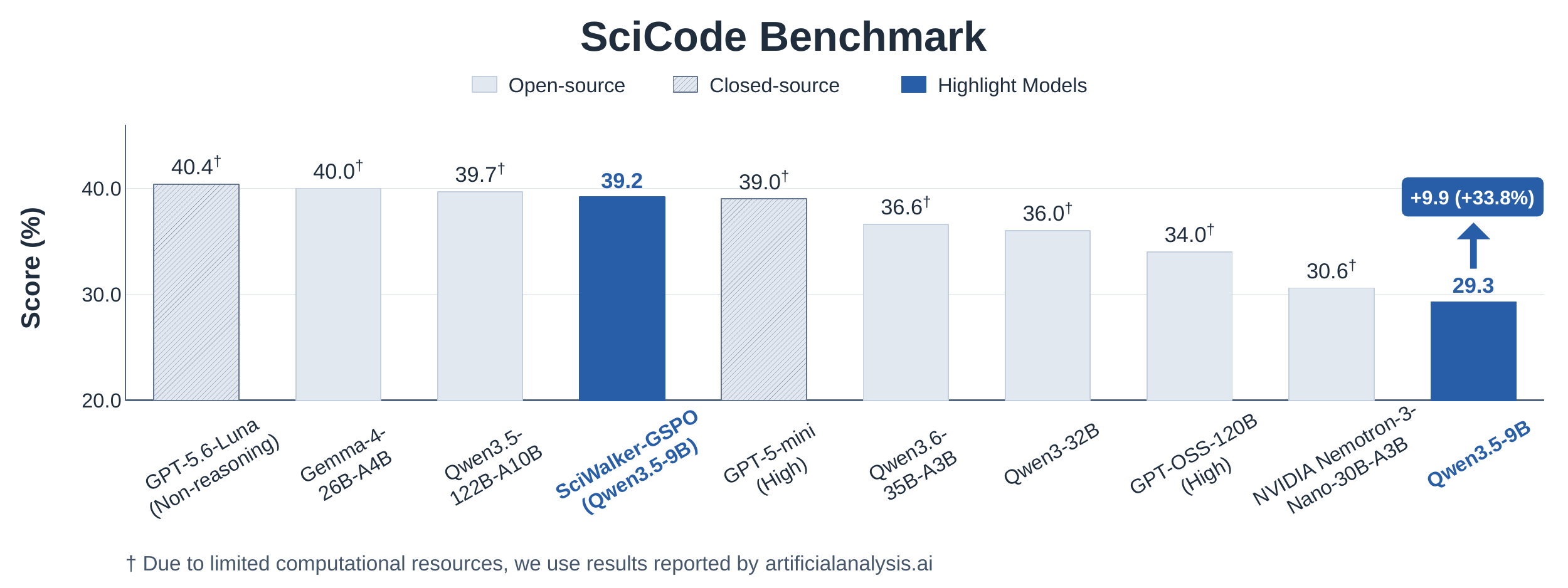}
\caption{SciCode performance comparison. Local evaluations report mean subproblem accuracies over three evaluations of 288 subproblems with scientific background. External model scores are reported by Artificial Analysis~\citep{artificialanalysis2026scicode}.}
\label{fig:scicode_comparison}
\end{figure}

\textbf{SciCode Performance.} 
Figure~\ref{fig:scicode_comparison} reports model performance on SciCode, combining our local evaluations with externally reported results. Each local evaluation covers 288 test-set subproblems with scientific background. By applying GSPO with SciWalker-generated training data, Qwen3.5-9B improves from 29.3\% to \textbf{39.2\%, an absolute gain of 9.9 points} (33.8\% relative). It surpasses substantially larger models such as Qwen3-32B (36.0\%) and GPT-OSS-120B (34.0\%), while matching GPT-5 mini (39.0\%) and approaching Qwen3.5-122B (39.7\%). These results demonstrate the effectiveness of SciWalker-generated data in substantially improving scientific reasoning capabilities, particularly for compact models.

\begin{table}[t]
\centering
\caption{Performance of the Qwen3.5-9B baseline, PPO, and GSPO across 16 benchmarks. Scores (\%) are means over three evaluations, rounded to one decimal before calculating the parenthesized percentage-point changes relative to the baseline. Increases and decreases are shown in dark blue and dark red, respectively. SciCode reports subproblem accuracy, code benchmarks report pass@1, and reasoning and knowledge benchmarks report accuracy. Bold indicates the highest mean score in each column, including ties.}
\label{tab:cross_benchmark_results}
\begingroup
\small
\setlength{\tabcolsep}{2pt}
\renewcommand{\arraystretch}{1.23}
\renewcommand{\tabularxcolumn}[1]{m{#1}}

\definecolor{deltablue}{HTML}{174A7E}
\definecolor{deltared}{HTML}{8B1E1E}

\newcommand{\benchhead}[1]{{\fontsize{8.3}{9.6}\selectfont\shortstack{#1}}}
\newcommand{\up}[1]{\,{\scriptsize\textcolor{deltablue}{(+#1)}}}
\newcommand{\down}[1]{\,{\scriptsize\textcolor{deltared}{(-#1)}}}
\newcommand{\same}{\,{\scriptsize(0.0)}}

\begin{tabularx}{\linewidth}{
    @{}
    >{\raggedright\arraybackslash}m{0.175\linewidth}
    *{5}{>{\centering\arraybackslash}X}
    @{}
}
\toprule
& \multicolumn{5}{c}{
    \textcolor{sodablue}{
        \textbf{Scientific computing and code generation}
    }
} \\
\cmidrule(l){2-6}
\textbf{Model}
& \benchhead{SciCode}
& \benchhead{DS-1000}
& \benchhead{HumanEval}
& \benchhead{LCB\\Code Gen. v6}
& \benchhead{APPS\\Introductory} \\
\midrule
Baseline
& 29.3
& 45.6
& 94.7
& 61.9
& 69.1 \\

PPO
& 32.6\up{3.3}
& 49.4\up{3.8}
& 93.5\down{1.2}
& 57.6\down{4.3}
& 72.3\up{3.2} \\

\rowcolor{sodablue!6}
GSPO
& \textbf{39.2}\up{9.9}
& \textbf{62.9}\up{17.3}
& \textbf{97.8}\up{3.1}
& \textbf{74.9}\up{13.0}
& \textbf{84.6}\up{15.5} \\
\end{tabularx}

\par\vspace{7pt}

\begin{tabularx}{\linewidth}{
    @{}
    >{\raggedright\arraybackslash}m{0.175\linewidth}
    *{6}{>{\centering\arraybackslash}X}
    @{}
}
& \multicolumn{6}{c}{
    \textcolor{sodablue}{
        \textbf{Code repair and execution}
    }
} \\
\cmidrule(l){2-7}
\textbf{Model}
& \benchhead{HumanEvalFix\\Python}
& \benchhead{HumanEvalFix\\Rust}
& \benchhead{HumanEvalFix\\JavaScript}
& \benchhead{QuixBugs\\Java}
& \benchhead{QuixBugs\\Python}
& \benchhead{LCB\\Execution v2} \\
\midrule
Baseline
& 80.5
& 51.2
& 43.3
& 58.3
& 77.5
& 97.1 \\

PPO
& 82.7\up{2.2}
& \textbf{60.6}\up{9.4}
& 66.9\up{23.6}
& 58.3\same
& 63.3\down{14.2}
& \textbf{97.8}\up{0.7} \\

\rowcolor{sodablue!6}
GSPO
& \textbf{93.9}\up{13.4}
& 48.4\down{2.8}
& \textbf{91.1}\up{47.8}
& \textbf{65.0}\up{6.7}
& \textbf{87.5}\up{10.0}
& 97.4\up{0.3} \\
\end{tabularx}

\par\vspace{7pt}

\begin{tabularx}{\linewidth}{
    @{}
    >{\raggedright\arraybackslash}m{0.175\linewidth}
    *{5}{>{\centering\arraybackslash}X}
    @{}
}
& \multicolumn{5}{c}{
    \textcolor{sodablue}{
        \textbf{Reasoning and knowledge}
    }
} \\
\cmidrule(l){2-6}
\textbf{Model}
& \benchhead{MATH-500}
& \benchhead{GSM8K}
& \benchhead{BBH\\Arithmetic}
& \benchhead{ARC-Easy}
& \benchhead{MMLU-Pro\\CS} \\
\midrule
Baseline
& 86.1
& 91.1
& 92.9
& 98.9
& 84.3 \\

PPO
& 81.4\down{4.7}
& 90.4\down{0.7}
& 89.5\down{3.4}
& 98.8\down{0.1}
& \textbf{85.0}\up{0.7} \\

\rowcolor{sodablue!6}
GSPO
& \textbf{92.1}\up{6.0}
& \textbf{96.0}\up{4.9}
& \textbf{97.1}\up{4.2}
& \textbf{99.0}\up{0.1}
& 84.6\up{0.3} \\
\bottomrule
\end{tabularx}

\par\vspace{4pt}

\begin{minipage}{\linewidth}
\footnotesize
\color{black!60}
* LCB denotes LiveCodeBench. BBH Arithmetic denotes BBH Multistep
Arithmetic. HumanEvalFix uses the test version. LiveCodeBench Execution
contains 479 execution instances from 92 source problems.
\end{minipage}
\endgroup
\end{table}

\textbf{Cross-Benchmark Performance.} 
Table~\ref{tab:cross_benchmark_results} compares the Qwen3.5-9B baseline with PPO and GSPO across 16 benchmarks. Overall, GSPO achieves substantial improvements over the baseline across a broad range of benchmarks. The gain is particularly pronounced on SciCode, where GSPO improves subproblem accuracy by 9.9 percentage points, demonstrating the effectiveness of the scientific-coding data generated by SciWalker. This benefit generalizes to other coding tasks that GSPO improves DS-1000 by 17.3 points, LiveCodeBench Code Generation v6 by 13.0 points, and APPS Introductory by 15.5 points.

The improvements also extend to out-of-domain tasks that are not directly targeted by the SciWalker data. On code repair, GSPO improves HumanEvalFix-Python and HumanEvalFix-JavaScript by 13.4 and 47.8 points, respectively. It also gains 6.7 points on QuixBugs-Java and 10.0 points on QuixBugs-Python. Moreover, GSPO consistently improves all five reasoning and knowledge benchmarks, such as gains of 6.0 points on MATH-500 and 4.9 points on GSM8K. These improvements are notable because these benchmarks are not coding tasks, suggesting that learning from high-quality scientific-coding trajectories can strengthen more general reasoning capabilities. Overall, GSPO improves 15 of the 16 benchmarks and raises the average score from 72.6\% to 82.0\%, indicating both the quality of the SciWalker-generated data and its broad generalization value.

Although PPO is less effective and less consistent than GSPO, it still improves SciCode by 3.3 points and produces clear gains on several coding-related benchmarks, such as DS-1000 (+3.8), APPS Introductory (+3.2). These gains provide additional evidence that the SciWalker-generated data contain useful learning signals. However, PPO improves only eight benchmarks, while degrading seven, and increases the overall mean by only 1.2 points. Its regressions on LiveCodeBench Code Generation and most reasoning benchmarks suggest that PPO does not exploit these signals as reliably as GSPO. One possible explanation is that critic-based credit assignment introduces estimation errors that make optimization less stable and limit generalization, highlighting the importance of the training objective in fully realizing the value of the generated data.



\FloatBarrier

\section{Conclusion}
\label{sec:7}

In this study, we introduced SciWalker, a framework for synthesizing scientific coding problems through operator-chain sampling and execution feedback. By using sampled operator chains as computational workflow cues, SciWalker generates scientifically grounded and diverse problems and iteratively repairs failed generations. Using this framework, we constructed 8,178 high-quality problems spanning 5 scientific domains and 32 subdomains. We conducted reinforcement learning on Qwen3.5-9B using GSPO algorithm and the training improved SciCode subproblem accuracy by 9.9 percentage points, from 29.3\% to 39.2\%, showing the effectiveness of the generated coding data. We also show that human-authored open-source code, such as Python libraries, is a valuable resource for synthetic coding data generation.

\FloatBarrier

\section*{Acknowledgments}

This work was supported by the Shanghai Artificial Intelligence Laboratory. We are grateful to the authors and open-source communities whose work made this project possible.

\begingroup
\small
\setlength{\bibsep}{3pt}
\bibliographystyle{iclr2026_conference}
\bibliography{paper}
\endgroup

\clearpage
\appendix
\section{From an \texorpdfstring{\texttt{fsolve}}{fsolve} Operator Chain to a Scientific Training Problem}
\label{app:a}

Using the actual problem \texttt{nonlinear\_parameter\_estimation}, this appendix illustrates how SciWalker converts an operator chain into a training problem.

\subsection{Operators and Chain Sampling}
\label{sec:A.1}

This example comes from the category of root finding and nonlinear equation solving, which is associated with SciPy interfaces such as \texttt{root}, \texttt{fsolve}, \texttt{brentq}, and \texttt{newton}. Among these, \texttt{scipy.optimize.fsolve} is combined with 35 operation modes to form 35 operators.

The sampled chain contains 11 operators and 10 edges, covering operations such as example adaptation, sensitivity analysis, data filtering, solving, residual computation, and stability checking. 

For example,

\begin{Verbatim}[breaklines,breakanywhere,fontsize=\footnotesize,frame=single,rulecolor=\color{sodarule},framesep=5pt]
fsolve x single_case_execute -> fsolve x compute_residual
\end{Verbatim}

This connection provides the computational cue check residuals after solving.

\subsection{From Scientific Context to a Concrete Problem}
\label{sec:A.2}

We use DeepSeek-V4-Flash. The operator chain receives a problem-generation potential assessment score of 7, and its scientific context is accepted after the first round of selection and review. The framework then generates the main problem, five substeps, student, oracle, and tests in a single complete problem-generation call.

The final problem concerns concentration decay in a chemical reaction. Given observations of time and concentration, the task is to fit the exponential model

\begin{equation}
\label{eq:decay-model}
y(t)=A\exp(-kt)+B
\end{equation}

by estimating its parameters \(A\), \(k\), and \(B\). The task requires first fitting the parameters, then filtering outlier observations using a residual threshold and refitting, returning the final parameters and the condition number of the approximate Hessian \(J^{\mathsf T}J\).

Table~\ref{tab:fsolve_substeps} summarizes the five substeps of this problem.

\begin{table}[!htbp]
\centering
\caption{Substeps of the generated nonlinear parameter estimation problem.}
\label{tab:fsolve_substeps}
\small
\setlength{\tabcolsep}{4pt}
\renewcommand{\arraystretch}{1.17}
\begin{tabularx}{\linewidth}{@{}lX@{}}
\toprule
\textbf{Substep} & \textbf{Computational task} \\
\midrule
1 & Compute the residual vector and Jacobian matrix \\
2 & Compute one Gauss--Newton parameter update \\
3 & Solve for the parameters using Gauss--Newton iterations with backtracking \\
4 & Compute the condition number of the approximate Hessian \\
5 & Filter the data using an absolute residual threshold \\
\bottomrule
\end{tabularx}
\end{table}

\subsection{Execution Validation and Feedback-Driven Repair}
\label{sec:A.3}

After initial generation, the framework runs problem checks and tests, then feeds the errors and the previous complete problem back for repair. This example first passes full validation after three rounds of repair (Table~\ref{tab:fsolve_repair}).

\begin{table}[!htbp]
\centering
\caption{Execution validation and feedback-driven repair of the example problem.}
\label{tab:fsolve_repair}
\small
\setlength{\tabcolsep}{4pt}
\renewcommand{\arraystretch}{1.17}
\begin{tabularx}{\linewidth}{@{}>{\raggedright\arraybackslash}p{0.20\linewidth}X@{}}
\toprule
\textbf{Version} & \textbf{Main changes and validation results during generation} \\
\midrule
Initial generation & The problem statement directly hints at interfaces; parameter recovery, differential testing, and other checks fail \\
First repair & Backtracking is added to student and oracle, and one differential test case is modified; parameter recovery errors remain after outlier handling \\
Second repair & A lower-bound constraint on the decay rate \(k\) is added to both implementations; full validation still fails \\
Third repair & The normal equations are solved using a least-squares method, and the iteration stopping criterion is adjusted; full validation passes \\
\bottomrule
\end{tabularx}
\end{table}

Four test snippets compare the reference and oracle implementations on the same inputs. Their number remains unchanged throughout repair, with one snippet modified during the first repair. This example illustrates the joint revision of the problem statement, implementations, and tests.

\subsection{Quality Grading, Inclusion, and Revalidation}
\label{sec:A.4}

After the whole repair progress, the framework proceeds to problem statement polishing and quality grading. In this example, the polishing response is identical to the repaired content and passes revalidation. The problem ultimately passes model-based quality review with an overall score of 8 and is included as a high-quality problem. The entire generation process involves 9 large language model calls, covering potential assessment, context selection and review, initial problem generation, three rounds of repair, polishing, and quality review.

\FloatBarrier

\section{Stage-Wise Retention Rates During Generation}
\label{app:b}

We report the number of candidates retained and the retention rate at each stage of problem generation (Table~\ref{tab:retention}) to quantify the extent of filtering. All retention rates below are calculated relative to the initial 66,889 candidate operator chains.

\begin{table}[!htbp]
\centering
\caption{Candidate retention during SciWalker data generation. All rates use the initial 66,889 candidate operator chains as the denominator.}
\label{tab:retention}
\small
\setlength{\tabcolsep}{4pt}
\renewcommand{\arraystretch}{1.17}
\begin{tabularx}{\linewidth}{@{}Xrrr@{}}
\toprule
\textbf{Stage} & \textbf{Retained} & \textbf{Reduction} & \textbf{Cumulative retention} \\
\midrule
Candidate operator chains & 66,889 & --- & 100.0\% \\
Problem-generation potential assessment & 21,543 & 45,346 & 32.2\% \\
Scientific context selection and review & 21,527 & 16 & 32.2\% \\
Training problem generation and execution validation (including repair) & 17,166 & 4,361 & 25.7\% \\
Problem statement polishing and quality grading (retaining only high-quality problems) & 8,178 & 8,988 & 12.2\% \\
\bottomrule
\end{tabularx}
\end{table}

The largest reduction occurs during problem-generation potential assessment, where 45,346 candidates are discarded. Because connecting operators into a chain in the graph does not guarantee that they can naturally form a task with a scientific rationale. After this stage, 21,543 candidate chains proceed to subsequent stages, accounting for 32.2\% of the initial candidates.

During scientific context selection and review, the model designs a scientific setting for each candidate chain and reviews the suitability of the solution objective and combination of operations. 21,527 candidate chains proceed to training problem generation, a reduction of 16 from the previous stage. This result includes context reselection.

During training problem generation and execution validation, the model generates problem statements, reference implementations, and tests. The framework runs checks and feeds errors back to the model for repair. Ultimately, 17,166 problems pass execution validation, a reduction of 4,361 from the previous stage, accounting for 25.7\% of the initial candidates.

During problem statement polishing and quality grading, the model first polishes the problem statements based on the code and tests, then reviews the problems for scientific naturalness, task difficulty, implementation complexity, and test strength. Only 8,178 high-quality problems are retained, accounting for 12.2\% of the initial candidates.

\FloatBarrier

\section{Generation Behavior of the Baseline Model}
\label{app:c}

This appendix examines the generation behavior of the Qwen3.5-9B baseline.

\subsection{Task and Evaluation Settings}
\label{sec:C.1}

\textbf{SciCode 14.1: Problem Statement and Scientific Background.}

\textbf{Problem statement}

Implement a python function to employ Mannella's leapfrog method to solve the Langevin equation of a microsphere optically trapped in the gas with the given initial condition.

\textbf{Scientific background}

For a microsphere trapped in the gas, we have the following Langevin equation: 
\begin{equation}
\label{eq:langevin}
\frac{{{d^2}x}}{{d{t^2}}} + \frac{{dx}}{{dt}}/{\tau _p} + \omega _0^2x = \sqrt {\frac{2}{{{\tau _p}}}} {v_{rms}}\zeta (t),
\end{equation}
 where \(\omega_0\) is the resonant frequency of the optical trap, \(\tau_p\) is the momentum relaxation time of the particle, \(v_{rms}\) is the root mean square velocity of the particle and \(\zeta(t)\) is a normalized white-noise process. This stochastic differential equation can be rewritten as:

\begin{gather}
v = \frac{{dx}}{{dt}}
\label{eq:velocity-definition}\\
\frac{{dv}}{{dt}} =  - v/{\tau _p} - \omega _0^2x + \sqrt {\frac{2}{{{\tau _p}}}} {v_{rms}}\zeta (t)
\label{eq:velocity-dynamics}
\end{gather}

Mannella's leapfrog method with step-size \(\Delta t\) defined as:

\begin{gather}
{x_{n + 1/2}} = {x_n} + {v_n}\Delta t/2,
\label{eq:leapfrog-midpoint}\\
{v_{n + 1}} = ({v_n} - {v_n}\Delta t/(2{\tau _p}) - \omega _0^2{x_{n + 1/2}}\Delta t + \sqrt {\frac{2}{{{\tau _p}}}} {v_{rms}}\Delta W)/(1 + \Delta t/(2{\tau _p})),
\label{eq:leapfrog-velocity}\\
{x_{n + 1}} = {x_{n + 1/2}} + {v_{n + 1}}\Delta t/2,
\label{eq:leapfrog-position}
\end{gather}

where \(\Delta W\) is sampled from a Gaussian distribution with mean zero and standard deviation \(\sqrt{\Delta t}\).

\Needspace{10\baselineskip}

\textbf{Required dependency}

\begin{Verbatim}[breaklines,breakanywhere,fontsize=\footnotesize,frame=single,rulecolor=\color{sodarule},framesep=5pt]
import numpy as np
\end{Verbatim}

\Needspace{10\baselineskip}

\textbf{Function interface and input/output specification}

\begin{Verbatim}[breaklines,breakanywhere,fontsize=\footnotesize,frame=single,rulecolor=\color{sodarule},framesep=5pt]
def harmonic_mannella_leapfrog(x0, v0, t0, steps, taup, omega0, vrms):
    '''Function to employ Mannella's leapfrog method to solve the Langevin equation of a microsphere optically trapped in the gas.
    Input
    x0 : float
        Initial position of the microsphere.
    v0 : float
        Initial velocity of the microsphere.
    t0 : float
        Total simulation time.
    steps : int
        Number of integration steps.
    taup : float
        Momentum relaxation time of the trapped microsphere in the gas (often referred to as the particle relaxation time).
    omega0 : float
        Resonant frequency of the harmonic potential (optical trap).
    vrms : float
        Root mean square velocity of the trapped microsphere in the gas.
    Output
    x : float
        Final position of the microsphere after the simulation time.
    '''
\end{Verbatim}

The problem requires passing the updated position and velocity to the next integration iteration. The omitted velocity-state update examined below occurs at this point.

The baseline uses the Inspect evaluation protocol. Scientific background is provided, thinking is enabled, the generation limit is 32,768 tokens, temperature=0.6, top\_p=0.95, top\_k=20, and the subproblem execution timeout is 300 seconds. The baseline is evaluated independently three times.

\subsection{Results and Generation Lengths Across Three Evaluations}
\label{sec:C.2}

On this subproblem, the baseline passes none of the three evaluations (Table~\ref{tab:case_results}).

\begin{table}[!htbp]
\centering
\caption{SciCode 14.1: baseline results and generation lengths across three independent evaluations. Length includes both reasoning and the final answer.}
\label{tab:case_results}
\footnotesize
\setlength{\tabcolsep}{4pt}
\renewcommand{\arraystretch}{1.17}
\begin{tabularx}{\linewidth}{@{}lrX@{}}
\toprule
\textbf{Evaluation} & \textbf{Tokens} & \textbf{Result} \\
\midrule
First & 1,496 & Generates code but omits the velocity-state update; fails the tests \\
Second & 1,326 & Generates code but omits the velocity-state update; fails the tests \\
Third & 32,768 & Repeatedly discusses dependency imports, reaches the length limit, and produces no final code \\
\midrule
Mean / Total & 11,863.33 & 0/3 pass \\
\bottomrule
\end{tabularx}
\end{table}

Generation length is the total number of generated tokens, including thinking text and the final answer. One baseline response is truncated. None of the three evaluations encounters an execution timeout.

\subsection{Persistent Looping and State Updates}
\label{sec:C.3}

\textbf{Dependency import loop.} In the third evaluation, the baseline repeatedly alternates between import NumPy for execution and follow the evaluation prompt and do not redeclare dependencies, without making further progress on the integration implementation. After removing blank lines and leading and trailing whitespace, the longest consecutive sequence of complete repetitions of the same four-line unit spans 308 iterations. This continuous region accounts for 82.11\% of the thinking text (by character count). Generation eventually exhausts the 32,768-token budget and terminates with \texttt{max\_tokens}, leaving the final code empty.

\textbf{Velocity-state update.} In the first and second evaluations, the baseline generates code but, after computing \texttt{v\_new}, uses it only to update the position without updating the velocity variable \texttt{v} to \texttt{v\_new}. The next iteration therefore continues to read the old velocity, violating the stepwise update requirement in Section~\ref{sec:C.1}.

\FloatBarrier

\section{Rollout Probability Correction and Numerical Details}
\label{app:d}

This appendix describes the additional rollout probability correction and numerical safeguards used with the GSPO-token formulation in Section~\ref{sec:4.1}. Let \(m_{i,k}\) mark valid generated tokens. We denote current-policy token log probabilities by \(\ell_{i,k}^{\theta}\), old-policy log probabilities recomputed on the training side by \(\ell_{i,k}^{\mathrm{old}}\), and log probabilities recorded during rollout by \(\ell_{i,k}^{\mathrm{roll}}\). All probabilities are conditioned on the sample's actual prompt and preceding response tokens.

The token-level rollout correction weight is
\begin{equation}
\label{eq:rollout-correction}
c_{i,k}=\operatorname{sg}\!\left[
\min\!\left(2,\ \exp\!\left[
\operatorname{clip}(\ell_{i,k}^{\mathrm{old}}-\ell_{i,k}^{\mathrm{roll}},-20,20)
\right]\right)\right].
\end{equation}
These weights are computed and fixed before the policy update. We do not normalize them across the batch or reject trajectories based on these weights. They multiply the token losses without modifying rewards or group-relative advantages.

For numerical stability, Equation~\ref{eq:token-ratio} is evaluated in log space with its exponent capped at 10. Writing \(\overline\Delta_i=\log\rho_i(\theta)\), the resulting ratio is
\begin{equation}
\label{eq:stabilized-token-ratio}
\widetilde\rho_{i,k}=
\exp\!\left[\min\!\left(
\operatorname{sg}(\overline\Delta_i)
+\ell_{i,k}^{\theta}-\operatorname{sg}(\ell_{i,k}^{\theta}),\ 10
\right)\right].
\end{equation}

With \(\epsilon_{\mathrm{low}}=3\times10^{-4}\) and \(\epsilon_{\mathrm{high}}=4\times10^{-4}\), the implemented loss combines this ratio with the rollout correction weights:
\begin{equation}
\label{eq:policy-objective}
\mathcal L(\theta)
=-\frac{1}{|\mathcal B|}\sum_{i\in\mathcal B}
\frac{\sum_k m_{i,k}\,c_{i,k}\,
\min\!\left[\widetilde\rho_{i,k}A_i,\,
\operatorname{clip}(\widetilde\rho_{i,k},1-\epsilon_{\mathrm{low}},1+\epsilon_{\mathrm{high}})A_i\right]}
{\sum_k m_{i,k}+10^{-8}}.
\end{equation}
The loss is averaged over valid tokens within each response and then over substep samples. Both reasoning and final code are included. Prompt and padding tokens are excluded. We use \(\delta=10^{-6}\) in Equation~\ref{eq:grpo-advantage}, with no Critic, KL loss, or entropy regularization.

\section{Scientific Python Libraries}
\label{app:e}

The operator catalogs draw on APIs from 79 Python libraries and package namespace groups across five scientific domains. Table~\ref{tab:scientific_libraries} lists these libraries by domain, with shared libraries appearing in multiple rows. Their APIs provide computational workflow cues for problem generation.

\begin{table}[!htbp]
\centering
\caption{Scientific libraries and package namespace groups used for operator construction, grouped by domain. Library names are alphabetized within each row.}
\label{tab:scientific_libraries}
\small
\setlength{\tabcolsep}{6pt}
\renewcommand{\arraystretch}{1.2}
\begin{tabularx}{\linewidth}{@{}>{\raggedright\arraybackslash}p{0.18\linewidth}X@{}}
\toprule
\textbf{Domain} & \textbf{Libraries} \\
\midrule
Mathematics & arch, ArviZ, CVXPY, geomdl, JAXopt, NLopt, NumPy, PyGMO, PyLops, PyMC, pymoo, Pyomo, PyVista, Riskfolio-Lib, scikit-learn, SciPy, SfePy, statsmodels, trimesh \\
\addlinespace[5pt]
Physics & alchemlyb, ASE, Astropy, Awkward Array, boost-histogram, Cirq, coffea, DecayLanguage, Diffractio, discretize, FiPy, Gala, galpy, GWpy, HCIPy, healpy, hist, Lightkurve, mplhep, NumPy, particle, POPPY, py\_pol, PyDMD, PySINDy, Qiskit, QuTiP, qutip-qip, REBOUND, SciPy, sisl, Stim, SymPy, zfit \\
\addlinespace[5pt]
Chemistry & Biopython, Cantera, datamol, DeepChem, edlib, molmass, MolVS, OpenMM, parasail, periodictable, pysam, PySCF, RDKit, scikit-bio, SciPy \\
\addlinespace[5pt]
Biology & agentpy, Biopython, COBRApy, DendroPy, edlib, GillesPy2, Mesa, msprime, NumPy, parasail, PyMC, pysam, scikit-bio, SciPy, tskit \\
\addlinespace[5pt]
Materials science & CHGNet, Dans\_Diffraction, datamol, jarvis-tools, jobflow, matminer, molmass, periodictable, pyFAI, pymatgen, RDKit \\
\bottomrule
\end{tabularx}
\end{table}

\end{document}